\documentclass[10pt,twocolumn,letterpaper]{article}

\usepackage{cvpr}
\usepackage{times}
\usepackage{epsfig}
\usepackage{graphicx}
\usepackage{amsmath}
\usepackage{amssymb}
\usepackage{algorithm}
\usepackage{algorithmicx}
\usepackage{algpseudocode}

\newcommand{\bigfigurewidth}{15.0cm}

\newcommand{\scatterwidth}{7.5cm}
\newcommand{\subfigurekern}{1.0cm}

\cvprfinalcopy 

\def\cvprPaperID{259} 

\begin{document}

\title{DisturbLabel: Regularizing CNN on the Loss Layer}

\author{Lingxi Xie\textsuperscript{1}
\thanks{This work was done when Lingxi Xie and Zhen Wei were interns at MSR.
This work was supported by ARO grant W911NF-15-1-0290,
Faculty Research Gift Awards by NEC Labs of America and Blippar, and NSFC 61322201, 61429201 and 61432019.
We thank Prof. Alan Yuille and the anonymous reviewers for instructive discussions.}\quad
Jingdong Wang\textsuperscript{2}\quad Zhen Wei\textsuperscript{3}\quad
Meng Wang\textsuperscript{4}\quad Qi Tian\textsuperscript{5}\\
\textsuperscript{1}Department of Statistics, University of California, Los Angeles, Los Angeles, CA, USA\\
\textsuperscript{2}Microsoft Research, Beijing, China\\
\textsuperscript{3}Department of Computer Science, Shanghai Jiao Tong University, Shanghai, China\\
\textsuperscript{4}School of Computer and Information, Hefei University of Technology, Hefei, Anhui, China\\
\textsuperscript{5}Department of Computer Science, University of Texas at San Antonio, San Antonio, TX, USA\\
\textsuperscript{1}{\tt\small 198808xc@gmail.com}\quad
\textsuperscript{2}{\tt\small jingdw@microsoft.com}\\
\textsuperscript{3}{\tt\small 94hazelnut@gmail.com}\quad
\textsuperscript{4}{\tt\small wangmeng@hfut.edu.cn}\quad
\textsuperscript{5}{\tt\small qitian@cs.utsa.edu}
}

\maketitle

\begin{abstract}
During a long period of time we are combating over-fitting in the CNN training process with model regularization,
including weight decay, model averaging, data augmentation, etc.
In this paper, we present {\bf DisturbLabel},
an extremely simple algorithm which randomly replaces a part of labels as incorrect values in each iteration.
Although it seems weird to intentionally generate incorrect training labels,
we show that DisturbLabel prevents the network training from over-fitting
by implicitly averaging over exponentially many networks which are trained with different label sets.
To the best of our knowledge, DisturbLabel serves as the first work which adds noises on the {\bf loss layer}.
Meanwhile, DisturbLabel cooperates well with Dropout to provide complementary regularization functions.
Experiments demonstrate competitive recognition results on several popular image recognition datasets.
\end{abstract}

\section{Introduction}
\label{Introduction}

Deep Convolutional Neural Networks (CNNs)~\cite{LeCun_1990_Handwritten}
have shown significant performance gains in image recognition~\cite{Krizhevsky_2012_ImageNet}.
The large image repository, ImageNet~\cite{Deng_2009_ImageNet},
and the high-performance computational resources such as GPUs played very important roles in the resurgence of CNNs.
Meanwhile, a number of research attempts on various aspects have been made to learn the deep hierarchical structure
better~\cite{Simonyan_2015_Very}\cite{Szegedy_2015_Going} and faster~\cite{Ioffe_2015_Batch}.
CNN also provides efficient visual features for
other tasks~\cite{Donahue_2014_DeCAF}\cite{Razavian_2014_CNN}\cite{Xia_2016_Pose}\cite{Xie_2015_Image}\cite{Xie_2016_InterActive}.

Many regularization techniques have been developed to prevent neural network from over-fitting,
{\em e.g.}, the $\ell_2$-regularization over the {\em weights} ({\em a.k.a.}, weight decay)~\cite{Krizhevsky_2009_Learning},
Dropout~\cite{Hinton_2012_Improving}
which discards randomly-selected {\em activations} on the (hidden) layers during training,
DropConnect~\cite{Wan_2013_Regularization} which sets randomly-selected {\em weights} to zero during training,
data argumentation which manipulates the {\em input data}~\cite{Ciresan_2010_Deep}\cite{Krizhevsky_2012_ImageNet},
and early stopping the iteration~\cite{Plaut_1986_Experiments}.

In this paper, we propose {\bf DisturbLabel} which imposes the regularization within the {\bf loss layer}.
In each training iteration, DisturbLabel randomly selects a small subset of samples (from those in the current mini-batch)
and randomly sets their ground-truth labels to be incorrect,
which results in a noisy loss function and, consequently, noisy gradient back-propagation.
To the best of our knowledge, this is the first attempt to regularize the CNN on the {\em loss layer}.
We show that DisturbLabel is an alternative approach to combining a large number of models that are trained with different noisy data.
Experimental results show that DisturbLabel achieves comparable performance with Dropout and that it,
in conjunction with Dropout, obtains better performance on several image classification benchmarks.

The rest of this paper is organized as follows.
Section~\ref{RelatedWork} briefly introduces related work.
The DisturbLabel algorithm is presented in Section~\ref{DisturbLabel}.
The discussions and the cooperation with Dropout are presented in Sections~\ref{Discussions} and~\ref{Cooperation}, respectively.
Experimental results are shown in Section~\ref{Experiments}, and we conclude our work in Section~\ref{Conclusions}.

\section{Related Work}
\label{RelatedWork}

\begin{table*}
\centering
\begin{tabular}{|l||c|c|c|c|c|c|}
\hline
Method & weight decay & Dropout      & DropConnect & data augmentation & stochastic pooling & {\bf DisturbLabel} \\
\hline
Units  & weights      & hidden nodes & weights     & input nodes       & pooling layer      & {\bf loss layer}   \\
\hline
\end{tabular}
\caption{
    Comparison with different CNN regularization techniques.
    Please refer to the texts for detailed references.
}
\label{Tab:ComparisingRegularizers}
\end{table*}

\begin{figure*}
\begin{center}
    \includegraphics[width=\bigfigurewidth]{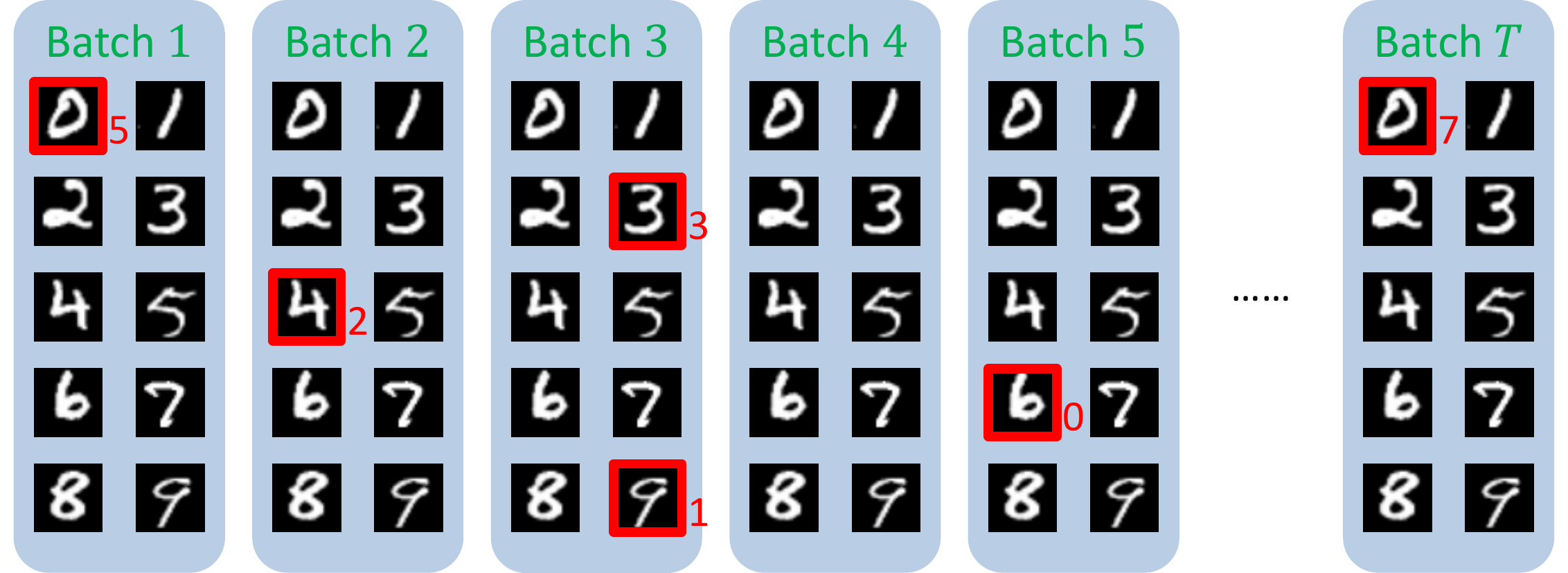}
\end{center}
\caption{
    An illustration of the DisturbLabel algorithm (${\alpha}={10\%}$).
    A mini-batch of $10$ training samples is used as the toy example.
    Each sample is disturbed with the probability $\alpha$.
    A disturbed training sample is marked with a red frame and the disturbed label is written below the frame.
    Even if a sample is disturbed, the label may remain unchanged ({\em e.g.}, the digit 3 in the $3$rd mini-batch).
}
\label{Fig:DisturbLabel}
\end{figure*}

The recent great success of CNNs in image recognition has benefitted from and inspired a wide range of research efforts,
such as designing deeper network structures~\cite{Simonyan_2015_Very}\cite{Szegedy_2015_Going},
exploring or learning non-linear activation functions~\cite{Goodfellow_2013_Maxout}\cite{Lin_2014_Network}\cite{He_2015_Delving},
developing new pooling operations~\cite{Zeiler_2013_Stochastic}\cite{Graham_2014_Fractional}\cite{Lee_2016_Generalizing},
introducing better optimization techniques~\cite{Lee_2015_Deeply},
regularization techniques preventing the network from over-fitting~\cite{Hinton_2012_Improving}\cite{Wan_2013_Regularization},
{\em etc}.

Avoiding over-fitting is a major challenge against training robust CNN models.
The early solutions include reducing the model complexity by using fewer parameters or sharing parameters~\cite{LeCun_1998_Gradient},
early stopping~\cite{Plaut_1986_Experiments} in which training is stopped before convergence,
weight decay~\cite{Krizhevsky_2009_Learning} which can be interpreted
as a way of constraining the parameters using the $\ell_2$-regularization and is now widely-adopted.

Recently, various regularization methods have been introduced.
Data augmentation generates more training data as the input of the CNN
by randomly cropping, rotating and flipping the {\em input images}~\cite{Krizhevsky_2012_ImageNet}\cite{Ciresan_2010_Deep},
and adding noises to the image {\em pixels}~\cite{Reed_1992_Regularization}.
Dropout~\cite{Hinton_2012_Improving} randomly discards a part of neuron response (the {\em hidden neurons})
during the training and only updates the remaining weights in each mini-batch iteration.
DropConnect~\cite{Wan_2013_Regularization} instead only updates a randomly-selected subset of {\em weights}.
Stochastic Pooling~\cite{Zeiler_2013_Stochastic} changes the deterministic {\em pooling operation}
and randomly samples one input as the pooling result in probability during training.
Probabilistic Maxout~\cite{Springenberg_2013_Improving} instead
turns the {\em Maxout operation}~\cite{Goodfellow_2013_Maxout} to stochastic.
In contrast, our approach (DisturbLabel) imposes the regularization at the {\em loss layer}.
A comparison of different regularization methods is summarized in Table~\ref{Tab:ComparisingRegularizers}.

There are other research works that are related to noisy labeling.
\cite{Sukhbaatar_2014_Training} explores the performance of discriminatively-trained CNNs on the noisy data,
where there are some freely available labels for each image which may or may not be accurate.
In contrast, our approach (DisturbLabel) assumes the labeling is correct
and randomly changes the labels in a small probability in each mini-batch iteration,
which means that the label of a training sample is correct in most iterations.
Different from other works adding noise to
the input unit~\cite{Vincent_2008_Extracting}\cite{Vincent_2010_Stacked}\cite{Maaten_2013_Learning}\cite{Maaten_2014_Marginalizing},
our work is a way of regularizing a neural network by adding noise to its loss layer (related to the output unit).

\section{The DisturbLabel Algorithm}
\label{DisturbLabel}

We start with the typical CNN training process.
The image dataset is given as ${\mathcal{D}}={\left\{\left(\mathbf{x}_n,\mathbf{y}_n\right)\right\}_{n=1}^N}$,
in which the data point is a $D$-dimensional vector ${\mathbf{x}_n}\in{\mathbb{R}^D}$,
and the label is a $C$-dimensional vector ${\mathbf{y}_n}={\left[0,\cdots,0,1,0,\cdots,0\right]^\top}$
with the entry of the corresponding class being $1$ and all the others being $0$.
The goal is to train a CNN model $\mathbb{M}$: ${\mathbf{f}\!\left(\mathbf{x};\boldsymbol{\theta}\right)}\in{\mathbb{R}^C}$,
in which $\boldsymbol{\theta}$ represents the model parameters.

$\boldsymbol{\theta}$ is often initialized as a set of white noises $\boldsymbol{\theta}_0$,
then updated using stochastic gradient descent (SGD)~\cite{Bottou_2010_Large}.
The $t$-th iteration of SGD updates the current parameters $\boldsymbol{\theta}_t$ as:
\begin{equation}
\label{Eqn:ModelUpdate}
{\boldsymbol{\theta}_{t+1}}=
    {\boldsymbol{\theta}_{t}+\gamma_t\cdot\frac{1}{|\mathcal{D}_t|}
        {\sum_{\left(\mathbf{x},\mathbf{y}\right)\in\mathcal{D}_t}}
        \nabla_{\boldsymbol{\theta}_t}\!\left[l\!\left(\mathbf{x},\mathbf{y}\right)\right]}.
\end{equation}
Here, $l\!\left(\mathbf{x},\mathbf{y}\right)$ is a loss function, {\em e.g.}, softmax or square loss.
$\nabla_{\boldsymbol{\theta}_t}\!\left[l\!\left(\mathbf{x},\mathbf{y}\right)\right]$ is computed using gradient back-propagation.
$\mathcal{D}_t$ is a mini-batch randomly drawn from the training dataset $\mathcal{D}$, and $\gamma_t$ is the learning rate.

{\bf DisturbLabel} works on each mini-batch independently.
It performs an extra sampling process, in which a disturbed label vector
${\widetilde{\mathbf{y}}}={\left[\widetilde{y}_1,\widetilde{y}_2,\ldots,\widetilde{y}_C\right]^{\top}}$
is randomly generated for each data $\left(\mathbf{x},\mathbf{y}\right)$
from a Multinoulli (generalized Bernoulli) distribution $\mathcal{P}\!\left(\alpha\right)$:
\begin{equation}
\label{Eqn:NoisyLabel}
\left\{\begin{array}{rcl}
{\widetilde{c}} & \sim & {\mathcal{P}\!\left(\alpha\right)}, \\
{\widetilde{y}_{\widetilde{c}}} & = & {1}, \\
{\widetilde{y}_{\widetilde{i}}} & = & {0},\quad\quad{\forall{i}\neq{\widetilde{c}}}.
\end{array}\right.
\end{equation}
The Multinoulli distribution $\mathcal{P}\!\left(\alpha\right)$
is defined as ${p_c}={1-\frac{C-1}{C}\cdot\alpha}$ and ${p_i}={\frac{1}{C}\cdot\alpha}$ for ${i}\neq{c}$.
$\alpha$ is the {\em noise rate}, and $c$ is the ground-truth label ({\em i.e.}, in the true label vector $\mathbf{y}$, ${y_c}={1}$).
In other words, we disturb each training sample with the probability $\alpha$.
For each disturbed sample, the label is randomly drawn from a uniform distribution over $\left\{1,2,\ldots,C\right\}$,
regardless of the true label.

\begin{algorithm}
\caption{DisturbLabel}
\begin{algorithmic}[1]
\State {\bf Input:} ${\mathcal{D}}={\left\{\left(\mathbf{x}_n,\mathbf{y}_n\right)\right\}_{n=1}^N}$, noise rate $\alpha$.
\State {\bf Initialization:} a network model $\mathbb{M}$:
${\mathbf{f}\!\left(\mathbf{x};\boldsymbol{\theta}_0\right)}\in{\mathbb{R}^C}$;
\For {each mini-batch ${\mathcal{D}_t}={\left\{\left(\mathbf{x}_m,\mathbf{y}_m\right)\right\}_{m=1}^M}$}{}
    \For {each sample $\left(\mathbf{x}_m,\mathbf{y}_m\right)$}{}
        \State Generate a disturbed label $\widetilde{\mathbf{y}}_m$ with Eqn~\eqref{Eqn:NoisyLabel};
    \EndFor
    \State Update the parameters $\boldsymbol{\theta}_t$ with Eqn~\eqref{Eqn:ModelUpdate};
\EndFor
\State {\bf Output:} the trained model $\mathbb{M}'$:
${\mathbf{f}\!\left(\mathbf{x};\boldsymbol{\theta}_T\right)}\in{\mathbb{R}^C}$.
\end{algorithmic}
\label{Alg:DisturbLabel}
\end{algorithm}

The pseudo codes of DisturbLabel are listed above.
An illustration of DisturbLabel is shown in Figure~\ref{Fig:DisturbLabel}.

\subsection{The Effect of the Noise Rate}
\label{DisturbLabel:NoiseRate}

The noise rate $\alpha$ determines the expected fraction ($\frac{C-1}{C}\cdot\alpha$)
of training data in a mini-batch which are assigned incorrect labels.
When ${\alpha}={0\%}$, there are no noises involved, and the algorithm degenerates to the ordinary case.
When ${\alpha}\rightarrow{100\%}$, we are actually discarding most of the labels
and the training process becomes nearly unsupervised (as the probability assigned to any class is nearly $\frac{1}{C}$).
It is often necessary to set a relatively small $\alpha$,
although it is possible to obtain an efficient network with a rather large $\alpha$
({\em e.g.}, training the {\bf LeNet} with ${\alpha}={90\%}$ achieves $<2\%$ testing error rate on {\bf MNIST}).

\begin{figure*}
\begin{center}
\begin{minipage}{\scatterwidth}
\centering
    \includegraphics[width=\scatterwidth]{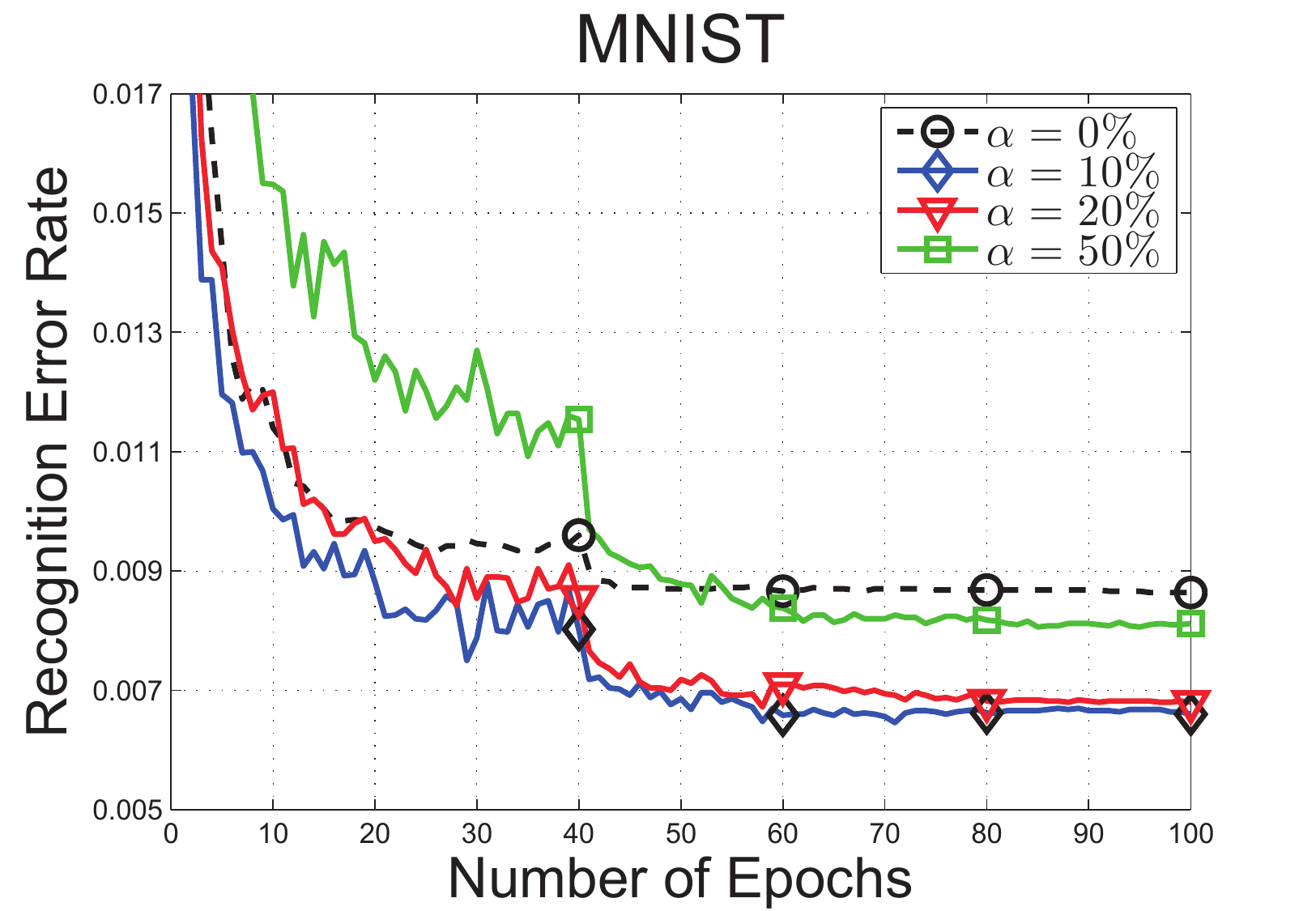}
\caption{
    {\bf MNIST} recognition error rate with respect to the noise level $\alpha$ (using {\bf LeNet}).
}
\label{Fig:MNISTNoise}
\end{minipage}
\hspace{\subfigurekern}
\begin{minipage}{\scatterwidth}
\centering
    \includegraphics[width=\scatterwidth]{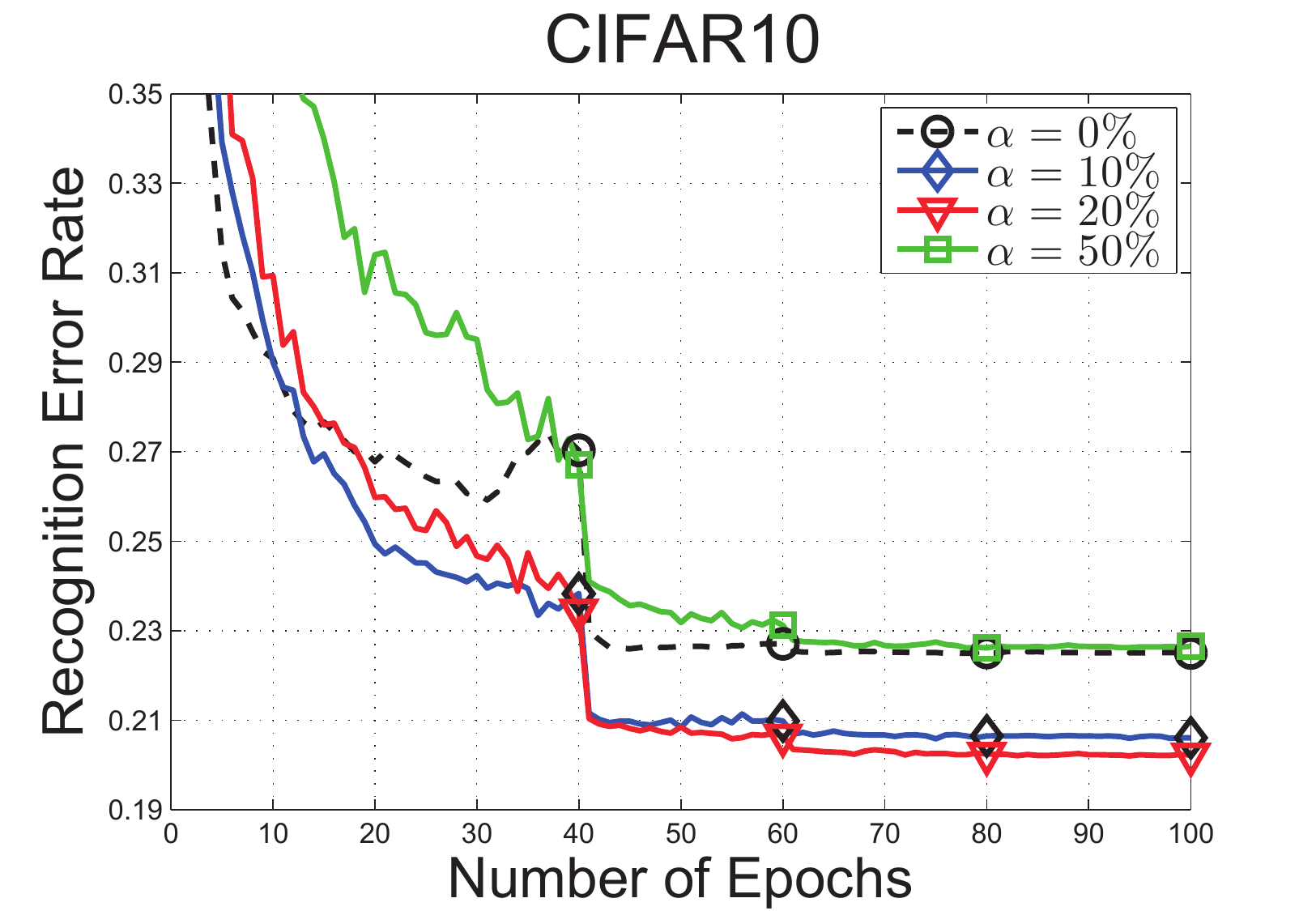}
\caption{
    {\bf CIFAR10} recognition error rate with respect to the noise level $\alpha$ (using {\bf LeNet}).
}
\label{Fig:CIFAR10Noise}
\end{minipage}
\end{center}
\end{figure*}

We evaluate the recognition accuracy on both {\bf MNIST} and {\bf CIFAR10},
by training the {\bf LeNet} with different noise rates $\alpha$,
We summarize the results in Figures~\ref{Fig:MNISTNoise} and~\ref{Fig:CIFAR10Noise}, respectively.
It is observed that DisturbLabel with a relatively small $\alpha$ (e.g., $10\%$ or $20\%$)
achieves higher recognition accuracy than the model without regularization (${\alpha}={0\%}$).
This verifies that DisturbLabel does improve the generalization ability of the trained CNN model
(Section~\ref{DisturbLabel:Regularization} provides an empirical verification
that the improvement comes from preventing over-fitting).
When the noise rate $\alpha$ goes up to $50\%$, DisturbLabel significantly causes the network to converge slower,
meanwhile produces lower recognition accuracy compared to smaller noises,
which is reasonable as the labels of the training data are not reliable enough.

\subsection{DisturbLabel as a Regularizer}
\label{DisturbLabel:Regularization}

\begin{figure*}
\begin{center}
\begin{minipage}{\scatterwidth}
\centering
    \includegraphics[width=\scatterwidth]{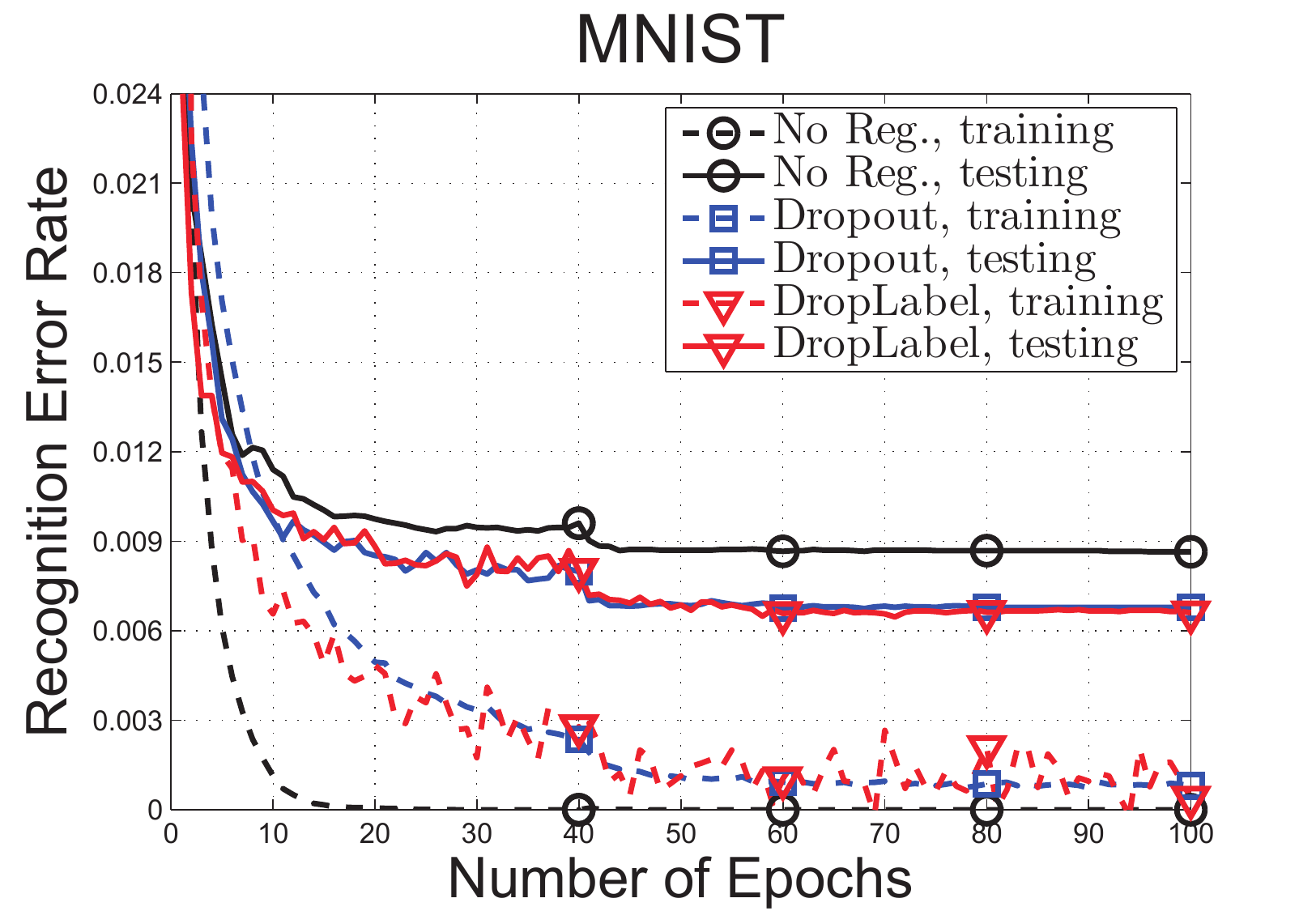}
\caption{
    {\bf MNIST} training vs. testing error on the {\bf LeNet} with: no regularization, Dropout and DisturbLabel.
}
\label{Fig:MNISTCurve}
\end{minipage}
\hspace{\subfigurekern}
\begin{minipage}{\scatterwidth}
\centering
    \includegraphics[width=\scatterwidth]{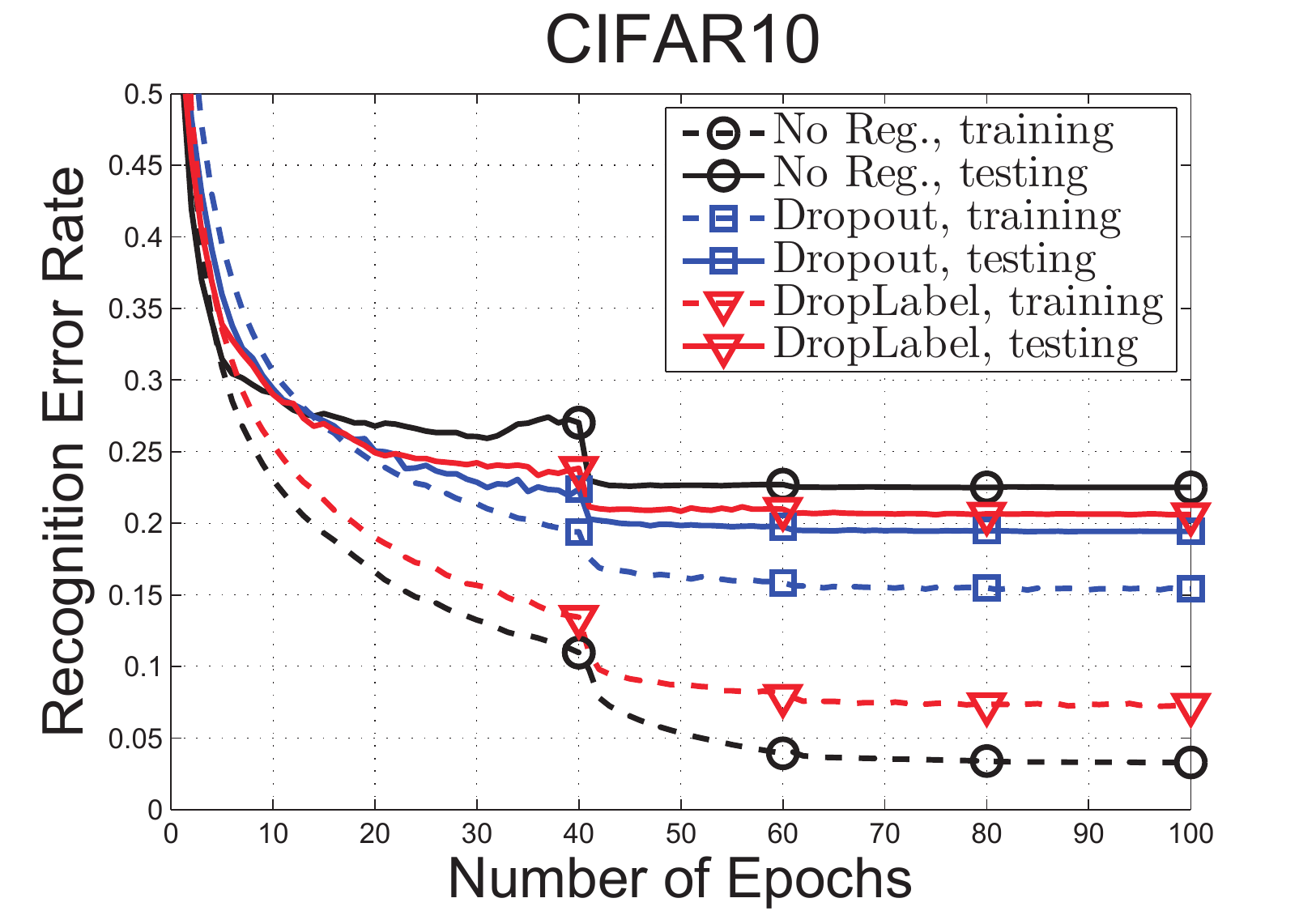}
\caption{
    {\bf CIFAR10} training vs. testing error on the {\bf LeNet} with: no regularization, Dropout and DisturbLabel.
}
\label{Fig:CIFAR10Curve}
\end{minipage}
\end{center}
\end{figure*}

We empirically show that DisturbLabel is able to prevent the network training from over-fitting.
Figures~\ref{Fig:MNISTCurve} and~\ref{Fig:CIFAR10Curve} show the results on the {\bf MNIST} and {\bf CIFAR10} datasets
using the normal training without regularization and the DisturbLabel algorithm over the same CNN structure.
In addition, we also report the results when training the CNN with Dropout which is an alternative approach of CNN regularization.

We can observe that without regularization, the training error quickly drops to quite a low level,
{\em e.g.}, almost $0\%$ on {\bf MNIST} and close to $3\%$ on {\bf CIFAR10}, but the testing error stops decreasing at a high level.
In contrast, the training error with DisturbLabel drops slower and is consistently larger than that without regularization.
However, the testing error becomes lower, verifying that the improvement comes from preventing over-fitting.
Similar results are also obtained in the case with the Dropout regularization.
This reveals that training a CNN with regularization (either DisturbLabel or Dropout) yields stronger generalization ability.

Theoretically, we apply DisturbLabel to a convex model.
Consider a linear regression task on the dataset $\mathcal{D}$, the linear model ${y}={\mathbf{w}^\top\mathbf{x}}$,
and the loss function ${L}={\frac{1}{2}{\sum_{n=1}^N}\left|{\mathbf{w}^\top\mathbf{x}_n} - y_n\right|^2}$.
The gradient over $\mathbf{w}$ is
${\frac{\partial L}{\partial\mathbf{w}}}={{\sum_{n=1}^N}\left(\mathbf{w}^\top\mathbf{x}_n-y_n\right)\mathbf{x}_n}$.
In our approach, each label $y_n$ may be disturbed as $\widetilde{y}_n$, and the gradient becomes
${\frac{\partial L}{\partial\mathbf{w}}}={{\sum_{n=1}^N}\left(\mathbf{w}^\top\mathbf{x}_n-\widetilde{y}_n\right)\mathbf{x}_n}$.
Their difference is ${\sum_{n=1}^N}(\widetilde{y}_n-y_n)\mathbf{x}_n$ assuming $\mathbf{w}$ is the same.
With $\ell_2$-regularization, the extra term $\frac{\lambda}{2}\left\|\mathbf{w}\right\|_2^2$ is added to the loss function,
and the term $\lambda\mathbf{w}$ added to the gradient over $\mathbf{w}$.
It is different from our approach which adds ${\sum_{n=1}^N}\left(\widetilde{y}_n-y_n\right)\mathbf{x}_n$.
Therefore, our approach, observed from the simple convex problem,
has a {\em damping effect}, but different from $\ell_2$-regularization.

\section{Discussions}
\label{Discussions}

\subsection{Difference from Soft Labeling}
\label{Discussions:SoftLabeling}

\begin{figure*}
\begin{center}
\begin{minipage}{\scatterwidth}
\centering
    \includegraphics[width=\scatterwidth]{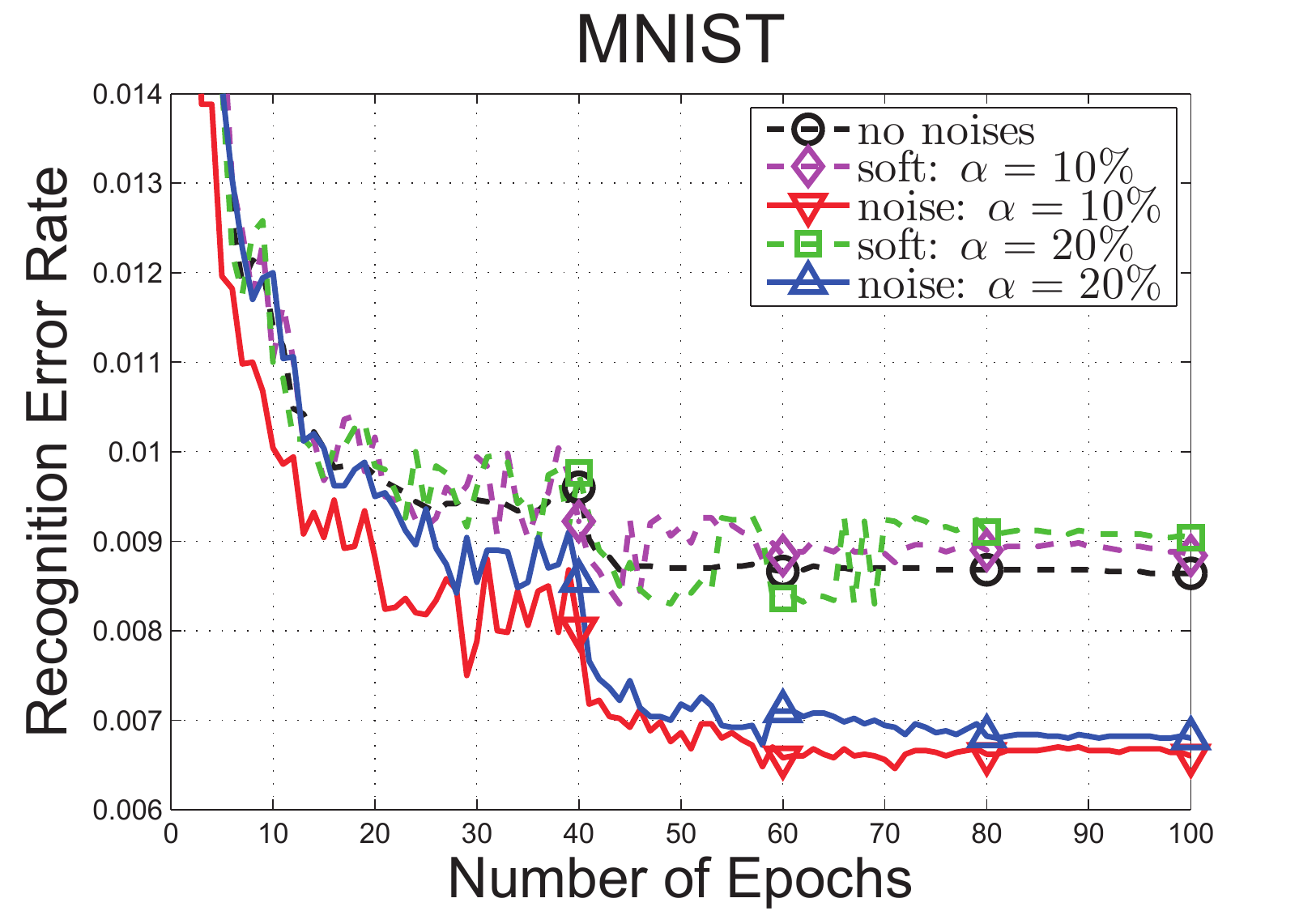}
\caption{
    {\bf MNIST} recognition error rate with soft labels and noisy labels (using the {\bf LeNet}).
}
\label{Fig:MNISTSoft}
\end{minipage}
\hspace{\subfigurekern}
\begin{minipage}{\scatterwidth}
\centering
    \includegraphics[width=\scatterwidth]{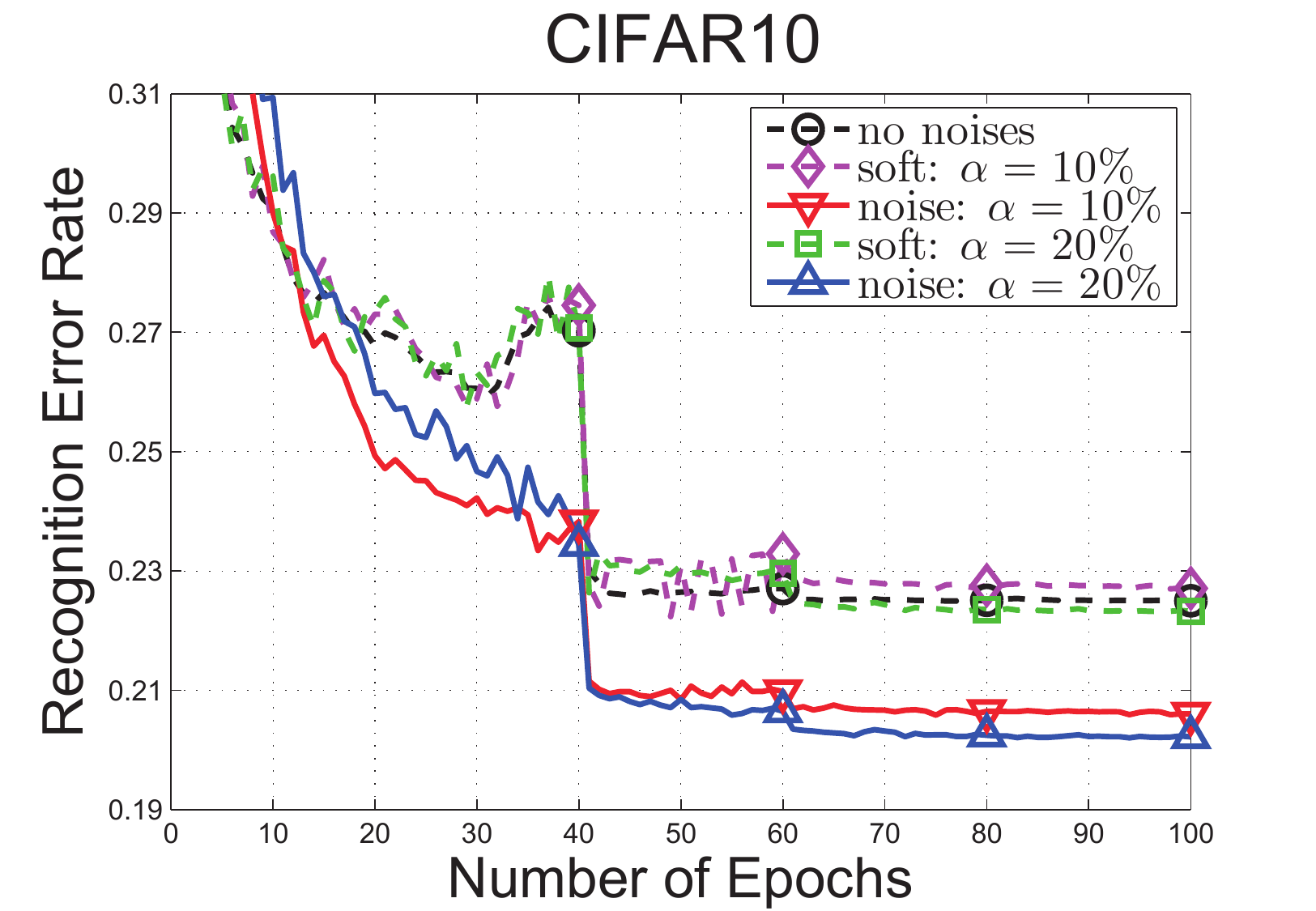}
\caption{
    {\bf CIFAR10} recognition error rate with soft labels and noisy labels (using the {\bf LeNet}).
}
\label{Fig:CIFAR10Soft}
\end{minipage}
\end{center}
\end{figure*}

Consider the soft labeling problem, where each data point $\mathbf{x}$ is assigned to a label with probability.
Denote the label vector by $\mathbf{y}'$, which is a $C$-dimensional vector,
with the $c$-th dimension being $1-\frac{C-1}{C}\cdot\alpha$ and all others being $\frac{\alpha}{C}$,
and $\alpha$ is the noise level.
Then we can normally train the CNN over the same data $\mathbf{x}$ and the soft label $\mathbf{y}'$,
using which the loss function is changed.
Here we consider the standard cross-entropy loss function,
and the derivation of other choices ({\em e.g.}, logistic loss, square loss, {\em etc.}) is similar.

The gradient for a training sample $\left(\mathbf{x},\mathbf{y}'\right)$ is:
\begin{equation}
\label{Eqn:SoftGradients}
{\frac{\partial}{\partial\boldsymbol{\theta}}\left|\mathbf{y}_n'-\mathbf{f}\right|_1}=
    {-\left[\frac{\partial\left|\mathbf{y}_n'-\mathbf{f}\right|_1}
        {\partial\left(\mathbf{y}_n'-\mathbf{f}\right)}\right]^\top\cdot
        \frac{\partial\mathbf{f}}{\partial\boldsymbol{\theta}}}
\end{equation}
However in DisturbLabel, the gradient is computed as:
\begin{equation}
\label{Eqn:DisturbGradients}
{\frac{\partial}{\partial\boldsymbol{\theta}}\left|\widetilde{\mathbf{y}}_n-\mathbf{f}\right|_1}=
    {-\left[\frac{\partial\left|\widetilde{\mathbf{y}}_n-\mathbf{f}\right|_1}
        {\partial\left(\widetilde{\mathbf{y}}_n-\mathbf{f}\right)}\right]^\top\cdot
        \frac{\partial\mathbf{f}}{\partial\boldsymbol{\theta}}}
\end{equation}

The empirical evaluation on the {\bf MNIST} and {\bf CIFAR10} datasets
is shown in Figures~\ref{Fig:MNISTSoft} and~\ref{Fig:CIFAR10Soft}, respectively.
One can observe that using soft labels generates nearly the same accuracy compared to ordinary training,
whereas DisturbLabel significantly improves recognition accuracy.
It is not surprising that DisturbLabel is not equivalent to soft label though
the expectation of the gradient in DisturbLabel is equal to the gradient in soft label
(as ${\mathbb{E}\!\left(\widetilde{\mathbf{y}}\right)}={\mathbf{y}'}$).
This reveals that using soft labels does not bring the regularization ability as DisturbLabel does,
which is also validated in the distillation solution~\cite{Hinton_2014_Distilling},
similar to the soft labeling, though it brings an advantage, making the network converge faster.

\subsection{Interpretation as Model Ensemble}
\label{DisturbLabel:ModelEnsemble}

We show that DisturbLabel can be interpreted as an implicit model ensemble.
Consider a normal noisy dataset
${\widetilde{\mathcal{D}}}={\left\{\left(\mathbf{x}_n,\widetilde{\mathbf{y}}_n\right)_{n=1}^N\right\}}$,
which is generated by assigning an incorrect label ${\widetilde{\mathbf{y}}_n}\neq{\mathbf{y}_n}$ to $\mathbf{x}_n$
with a probability $\alpha$ for each data point in $\mathcal{D}$.
Combining neural network models that are trained on different noisy sets is usually helpful~\cite{Hinton_2014_Distilling}.
However, separately training nets is prohibitively expensive as there are exponentially many noisy datasets.
Even if we have already trained many different networks,
combining them at the testing stage is very costly and often infeasible.

Each iteration in the DisturbLabel training process is like an iteration when
the network is trained over a different noisy dataset $\widetilde{\mathcal{D}}$
where a mini-batch of samples $\widetilde{\mathcal{D}}_t$ are drawn.
Thus, training a neural network with DisturbLabel can be regarded as training many networks
with massive weight sharing but over different training data,
where each training sample is used very rarely.

It is interesting that Dropout can be interpreted as a way of approximately combining
exponentially many {\em different} neural network architectures trained on the {\em same} data efficiently,
while DisturbLabel can be regarded as a way of approximately combining
exponentially many neural networks with the {\em same} architecture but trained on {\em different} noisy data efficiently.
In Section~\ref{Cooperation}, we will show that DisturbLabel cooperates with Dropout to produce better results than individual models.

\subsection{Interpretation as Data Augmentation}
\label{Discussions:DataAugmentation}

We analyze DisturbLabel from a data augmentation perspective.
Considering a data point $\left(\mathbf{x},\mathbf{y}\right)$ and its incorrect label $\widetilde{\mathbf{y}}$,
the contribution to the loss is $\left|f\!\left(\mathbf{x}\right)-\widetilde{\mathbf{y}}\right|_1$.
This contribution can be rewritten as
$\left|\left(f\!\left(\mathbf{x}\right)-\widetilde{\mathbf{y}}+\mathbf{y}\right)-\mathbf{y}\right|_1$,
where ${\widetilde{f}\!\left(\mathbf{x}\right)}\doteq{f\!\left(\mathbf{x}\right)-\widetilde{\mathbf{y}}+\mathbf{y}}$
can be viewed as a noisy output.
Inspired by~\cite{Konda_2015_Dropout},
we can project the noisy output $\widetilde{f}\!\left(\mathbf{x}\right)$ back into the input space
by minimizing the squared error $\left\|\widetilde{f}\!\left(\mathbf{x}\right)-f\!\left(\widetilde{\mathbf{x}}\right)\right\|_2^2$,
where $\widetilde{\mathbf{x}}$ is the augmented sample.
In summary, the data point with a disturbed label $\left(\mathbf{x},\widetilde{\mathbf{y}}\right)$
can be transformed to an augmented data point $\left(\widetilde{\mathbf{x}},\mathbf{y}\right)$.

To verify that DisturbLabel indeed acts as data augmentation,
we evaluate the algorithm on the {\bf MNIST} dataset~\cite{LeCun_1998_Gradient}
with only $1\%$ ($600$) and $10\%$ ($6000$) training samples, meanwhile keep the total number of iterations unchanged,
{\em i.e.}, each training sample is used $100\times$ and $10\times$ times as it is used in the original training process.
With the {\bf LeNet}~\cite{LeCun_1990_Handwritten},
we obtain $10.92\%$ and $2.83\%$ error rates on the original testing set, respectively,
which are dramatic compared to $0.86\%$ when the network is trained on the complete set.
Meanwhile, in both cases, the training error rates quickly decrease to $0$,
implying that the limited training data cause over-fitting.
DisturbLabel significantly decreases the error rates to $6.38\%$ and $1.89\%$, respectively.
As a reference, the error rate on the complete training set is further decreased to $0.66\%$ by DisturbLabel.
This indicates that DisturbLabel improves the quality of network training with implicit data augmentation,
thus it serves as an effective algorithm especially in the case that the amount training data is limited.

\subsection{Relationship to Other CNN Training Methods}
\label{Discussions:Relationship}

\begin{figure*}
\begin{center}
\begin{minipage}{\scatterwidth}
\centering
    \includegraphics[width=\scatterwidth]{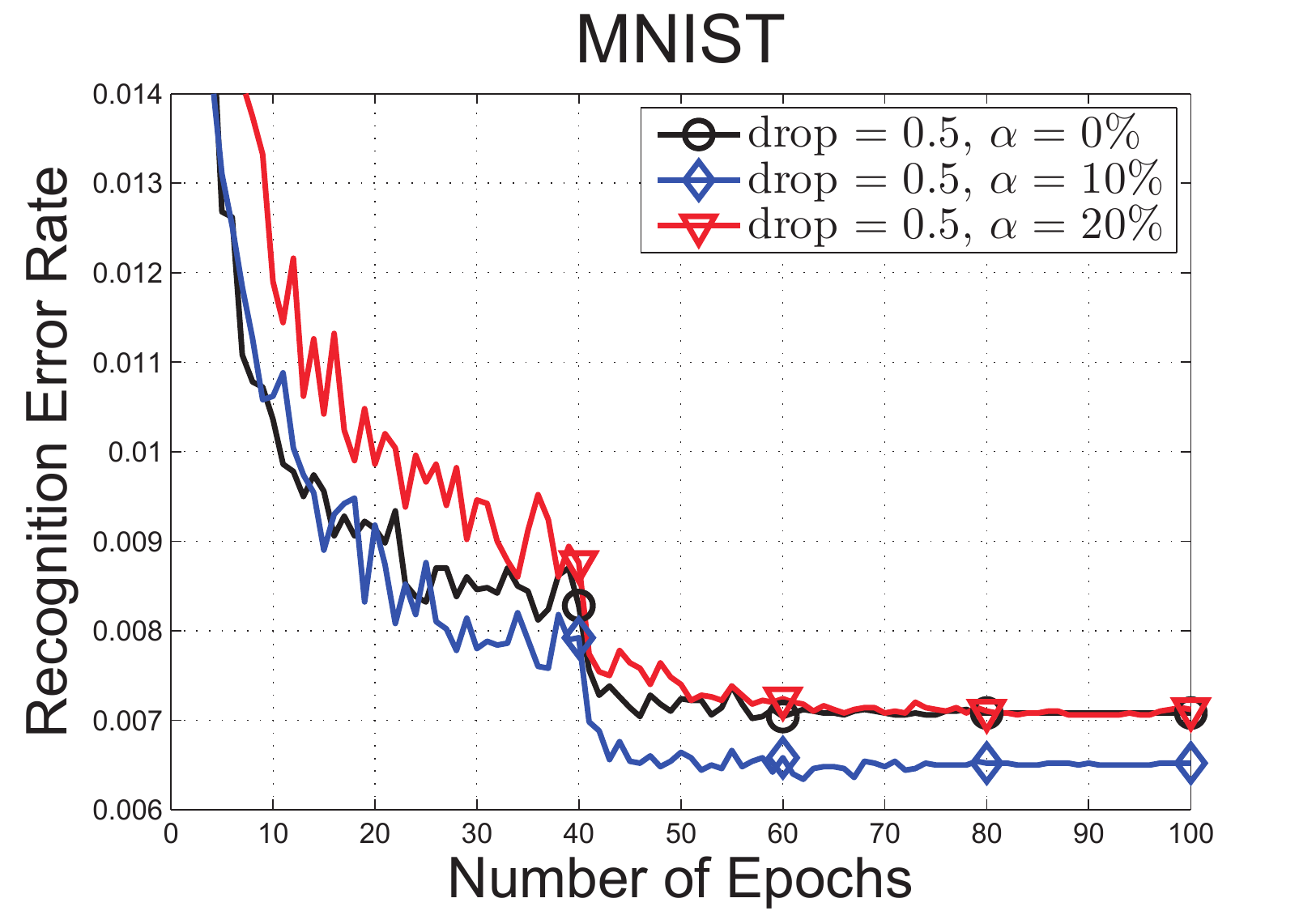}
\caption{
    {\bf MNIST} recognition error rate with drop rate $0.5$ and different noise levels $\alpha$ (using the {\bf LeNet}).
}
\label{Fig:MNISTDropout}
\end{minipage}
\hspace{\subfigurekern}
\begin{minipage}{\scatterwidth}
\centering
    \includegraphics[width=\scatterwidth]{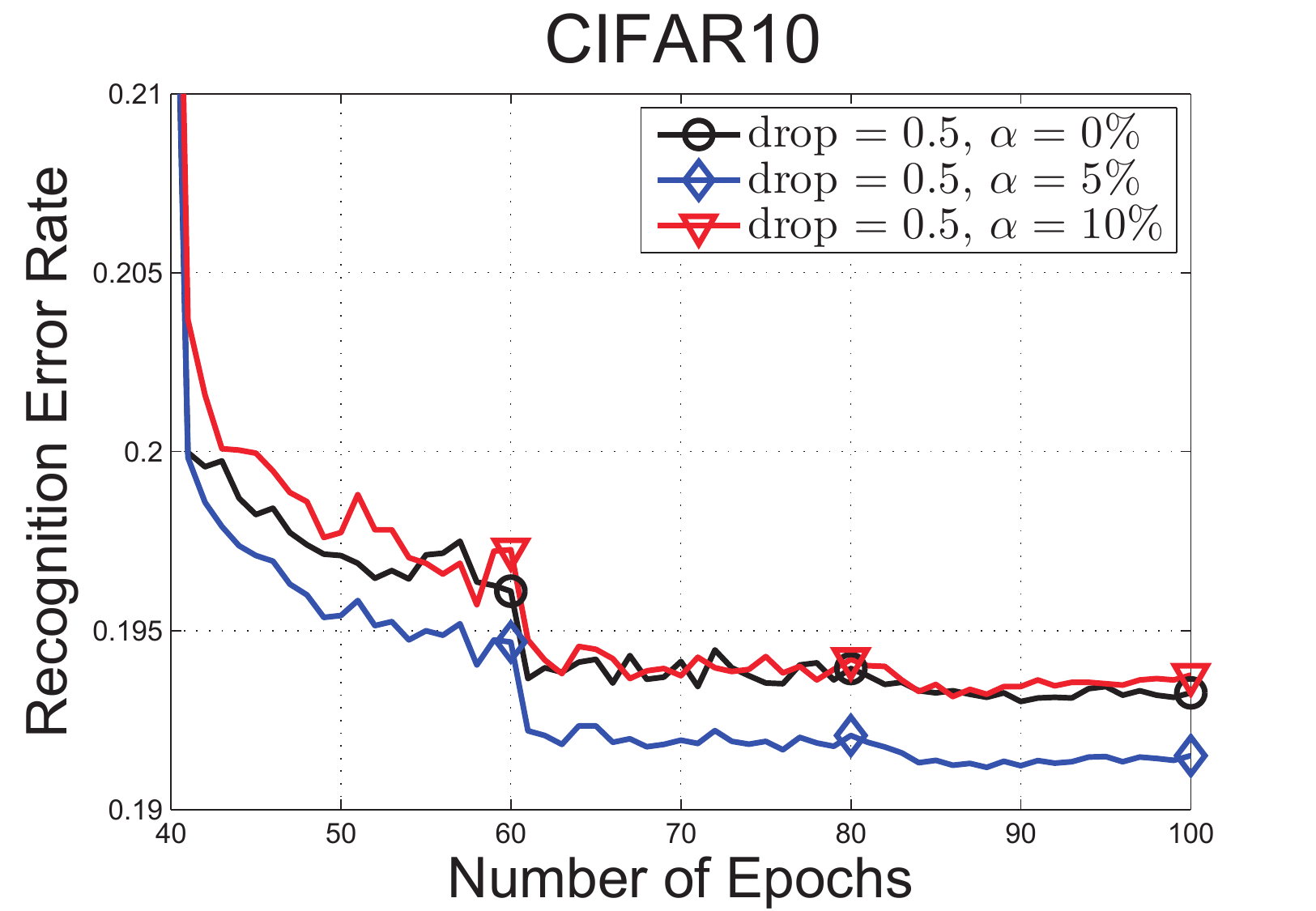}
\caption{
    {\bf CIFAR10} recognition error rate with drop rate $0.5$ and different noise levels $\alpha$ (using the {\bf LeNet}).
}
\label{Fig:CIFAR10Dropout}
\end{minipage}
\end{center}
\end{figure*}

We briefly discuss the relationship between DisturbLabel and other network training algorithms.

\begin{itemize}
\item {\bf Other regularization methods.}
There exist other network regularization methods, including DropConnect~\cite{Wan_2013_Regularization},
Stochastic Pooling~\cite{Zeiler_2013_Stochastic}, Probabilistic Maxout~\cite{Springenberg_2013_Improving}, {\em etc}.
Like Dropout~\cite{Hinton_2012_Improving} which regularizes CNNs on the hidden neurons,
these methods add regularization on other places such as neuron connections and pooling operations.
DisturbLabel regularizes CNNs on the loss layer, which, to the best of our knowledge, has never been studied before.
As we will show in the next section, DisturbLabel cooperates well with Dropout to obtain superior results to individual modules.
We believe that DisturbLabel can also provide complementary information to other network regularization methods.
\item {\bf Other methods dealing with noises.}
Some previous works~\cite{Hinton_2014_Distilling}\cite{Sukhbaatar_2014_Training} aim at training CNNs with noisy labels.
We emphasize that DisturbLabel is intrinsically different with these methods since the problem settings are completely different.
In these problems, training data suffer from noises (incorrect annotations),
and researchers discuss the possibility of overcoming the noises towards an accurate training process.
DisturbLabel, on the other hand, assumes that all the ground-truth labels are correct,
and intentionally generates incorrect labels on a small fraction of data to prevent the network from over-fitting.
In summary, inaccurate annotations may be harmful, but factitiously introducing noises is helpful to training a robust network.
\item {\bf Other network structures.}
We are also interested in other sophisticated network structures,
such as the Maxout Networks~\cite{Goodfellow_2013_Maxout}, the Deeply-Supervised Nets~\cite{Lee_2015_Deeply},
the Network-in-Network~\cite{Lin_2014_Network} and the Recurrent Neural Networks~\cite{Liang_2015_Recurrent}.
Thanks to the generality, DisturbLabel can be adopted on these networks to improve their generalization ability.
\end{itemize}

\section{Cooperation with Dropout}
\label{Cooperation}

We have shown that DisturbLabel regularizes the CNN on the loss layer.
This is different from Dropout~\cite{Hinton_2012_Improving}, which regularizes the CNN on hidden layers.
DisturbLabel is an approximate ensemble of many CNN models with the {\em same} structure trained over {\em different} noisy datasets,
while Dropout is an approximate ensemble of many CNN models with {\em different} structures trained over the {\em same} data.
These two regularization strategies are complementary.
We empirically discuss the combination of DisturbLabel with Dropout,
which leads to an ensemble of many CNN models with {\em different} structures trained over {\em different} noisy data.

We report the results with various noise levels $\alpha$ for DisturbLabel and a fixed drop rate for Dropout
over the {\bf MNIST} and {\bf CIFAR10} datasets in Figures~\ref{Fig:MNISTDropout} and~\ref{Fig:CIFAR10Dropout}, respectively.
In general, the proper combination of Dropout with DisturbLabel benefits the recognition accuracy improvement.
In the MNIST dataset, the best result is obtained when ${\alpha}={10\%}$, meanwhile ${\alpha}={20\%}$ performs much worse.
We note that in Figure~\ref{Fig:MNISTNoise}, without Dropout, ${\alpha}={10\%}$ and ${\alpha}={20\%}$ produce comparable results.
In the CIFAR10 dataset, ${\alpha}={10\%}$ does not help to improve the accuracy (close to baseline),
while in Figure~\ref{Fig:CIFAR10Noise}, we get comparable better results using ${\alpha}={10\%}$ and ${\alpha}={20\%}$.
The above experiments show that both DisturbLabel and Dropout add regularization to network training.
If both strategies are adopted, we need to reduce the regularization power properly to prevent ``under-fitting''.

In the later experiments, if both Dropout and DisturbLabel are used,
we will reduce the noise level $\alpha$ by half to prevent the regularization on network training from being too strong.
In the case that DisturbLabel provides strong regularization,
{\em e.g.}, in the case of {\bf ImageNet} training where the wrong label is distributed over all the $1000$ categories,
we will slightly decrease the drop rate of Dropout for the same purpose.

\section{Experiments}
\label{Experiments}

We evaluate DisturbLabel on five popular datasets,
{\em i.e.}, {\bf MNIST}~\cite{LeCun_1998_Gradient} and {\bf SVHN}~\cite{Netzer_2011_Reading} for digit recognition,
{\bf CIFAR10}/{\bf CIFAR100}~\cite{Krizhevsky_2009_Learning} for natural image recognition,
and {\bf ImageNet}~\cite{Deng_2009_ImageNet} for large-scale visual recognition.

\subsection{The MNIST Dataset}
\label{Experiments:MNIST}

\begin{table*}
\centering
\begin{tabular}{|l|r|r|||l|r|r|}
\hline
{\bf MNIST}                              & w/o DA & with DA & {\bf SVHN}                                  & w/o DA & with DA \\
\hline\hline
Jarrett~\cite{Jarrett_2009_What}         & $0.53$ & $-   $  & Sermanet~\cite{Sermanet_2012_Convolutional} & $5.15$ & $-   $  \\
Zeiler~\cite{Zeiler_2013_Stochastic}     & $0.47$ & $-   $  & Zeiler~\cite{Zeiler_2013_Stochastic}        & $2.80$ & $-   $  \\
Lin~\cite{Lin_2014_Network}              & $0.47$ & $-   $  & Goodfellow~\cite{Goodfellow_2013_Maxout}    & $2.47$ & $-   $  \\
Goodfellow~\cite{Goodfellow_2013_Maxout} & $0.45$ & $-   $  & Lin~\cite{Lin_2014_Network}                 & $2.35$ & $-   $  \\
Lee~\cite{Lee_2015_Deeply}               & $0.39$ & $-   $  & Wan~\cite{Wan_2013_Regularization}          & $-   $ & $1.94$  \\
Liang~\cite{Liang_2015_Recurrent}        & $0.31$ & $-   $  & Lee~\cite{Lee_2015_Deeply}                  & $1.92$ & $-   $  \\
Wan~\cite{Wan_2013_Regularization}       & $0.52$ & $0.21$  & Liang~\cite{Liang_2015_Recurrent}           & $1.77$ & $-   $  \\
\hline\hline
{\bf LeNet}, no Reg.                     & $0.86$ & $0.48$  & {\bf LeNet}, no Reg.                        & $3.93$ & $3.48$  \\
{\bf LeNet}, Dropout                     & $0.68$ & $0.43$  & {\bf LeNet}, Dropout                        & $3.65$ & $3.25$  \\
{\bf LeNet}, DisturbLabel                & $0.66$ & $0.45$  & {\bf LeNet}, DisturbLabel                   & $3.69$ & $3.27$  \\
{\bf LeNet}, both Reg.                   & $0.63$ & $0.41$  & {\bf LeNet}, both Reg.                      & $3.61$ & $3.21$  \\
\hline
{\bf BigNet}, no Reg.                    & $0.69$ & $0.39$  & {\bf BigNet}, no Reg.                       & $2.87$ & $2.35$  \\
{\bf BigNet}, Dropout                    & $0.36$ & $0.29$  & {\bf BigNet}, Dropout                       & $2.23$ & $2.08$  \\
{\bf BigNet}, DisturbLabel               & $0.38$ & $0.32$  & {\bf BigNet}, DisturbLabel                  & $2.28$ & $2.21$  \\
{\bf BigNet}, both Reg.                  & $0.33$ & $0.28$  & {\bf BigNet}, both Reg.                     & $2.19$ & $2.02$  \\
\hline
\end{tabular}
\caption{
    Recognition error rates ($\%$) on the {\bf MNIST} and {\bf SVHN} datasets.
    DA: data augmentation (random cropping).
}
\label{Tab:ComparisonDigit}
\end{table*}

{\bf MNIST}~\cite{LeCun_1998_Gradient} is one of the most popular datasets for handwritten digit recognition.
This dataset consists of $60000$ training images and $10000$ testing images, uniformly distributed over $10$ classes (0--9).
All the samples are $28\times28$ grayscale images.

We use a modified version of the {\bf LeNet}~\cite{LeCun_1990_Handwritten} as the baseline.
The input image is passed through two units consisting of convolution, ReLU and max-pooling operations.
In which, the convolutional kernels are of the scale $5\times5$, the spatial stride $1$,
and max-pooling operators are of the scale $2\times2$ and the spatial stride $2$.
The number of convolutional kernels are $32$ and $64$, respectively.
After the second max-pooling operation, a fully-connect layer with $512$ filters is added, followed by ReLU and Dropout.
The final layer is a $10$-way classifier with the softmax loss function.
We use a set of abbreviation to represent the above network configuration as:
[C$5$(S$1$P$0$)@$32$-MP$2$(S$2$)]-[C$5$(S$1$P$0$)@$64$-MP$2$(S$2$)]-FC$512$-D$0.5$-FC$10$.

To obtain higher recognition accuracy, we also train a more complicated {\bf BigNet}.
The cross-map normalization~\cite{Krizhevsky_2012_ImageNet} is adopted after each pooling layer,
and the parameter $K$ for normalization is proportional to the logarithm of the number of kernels.
The network configuration is abbreviated as:
[C$5$(S$1$P$2$)@$128$-MP$3$(S$2$)]-[C$3$(S$1$P$1$)@$128$-D$0.7$-C$3$(S$1$P$1$)@$256$-MP$3$(S$2$)]-D$0.6$-
[C$3$(S$1$P$1$)@$512$]-D$0.5$-[C$3$(S$1$P$1$)@$1024$-MP$S$(S$1$)]-D$0.4$-FC$10$.
Here, the number $S$ is the map size before the final (global) max-pooling, before which the down-sampling rate is $4$.
Therefore, if the input image size is $W\times W$, ${S}={\left\lfloor W/4\right\rfloor}$.
The {\bf BigNet} is feasible for data augmentation based on image cropping as the input image size is variable.

For data augmentation, we randomly crop input images into $24\times24$ pixels.
We apply $\left(40,20,20,20\right)$ training epochs for the {\bf LeNet}-based configurations
with learning rates $\left(10^{-3},10^{-4},10^{-5},10^{-6}\right)$.
For the {\bf BigNet}-based configurations,
the numbers are $\left(200,100,100,100\right)$ and $\left(10^{-2},10^{-3},10^{-4},10^{-5}\right)$, respectively.

We evaluate DisturbLabel with the noise level ${\alpha}={20\%}$.
According to the results shown in Table~\ref{Tab:ComparisonDigit},
DisturbLabel produces consistent accuracy gain over models without regularization,
and also cooperates with Dropout to further improve the recognition performance.
Train the {\bf BigNet} using both Dropout and DisturbLabel achieves a $0.33\%$ error rate without data augmentation,
which outperforms several recently reported results~\cite{Goodfellow_2013_Maxout}\cite{Lee_2015_Deeply}.
In comparison with~\cite{Ciresan_2010_Deep} which applies more complicated data augmentation ({\em e.g.}, image rotation),
we only use randomly image cropping and obtain a comparable error rate ($0.28\%$ vs. $0.23\%$).

\subsection{The SVHN Dataset}
\label{Experiments:SVHN}

The {\bf SVHN} dataset~\cite{Netzer_2011_Reading} is a larger collection of $32\times32$ RGB images,
{\em i.e.}, $73257$ training samples, $26032$ testing samples, and $531131$ extra training samples.
We preprocess the data as in the previous methods~\cite{Netzer_2011_Reading},
{\em i.e.}, selecting $400$ samples per category from the training set as well as $200$ samples per category from the extra set,
using these $6000$ images for validation, and the remaining $598388$ images as training samples.
We also use Local Contrast Normalization (LCN) for data preprocessing~\cite{Goodfellow_2013_Maxout}.

We use another version of the {\bf LeNet}.
A $32\times32\times3$ image is passed through three units consisting of convolution, ReLU and max-pooling operations.
Using abbreviation, the network configuration can be written as:
[C$5$(S$1$P$2$)@$32$-MP$3$(S$2$)]-[C$5$(S$1$P$2$)@$32$-MP$3$(S$2$)]-[C$5$(S$1$P$2$)@$64$-MP$3$(S$2$)]-FC$64$-D$0.5$-FC$10$.
Padding of $2$ pixels wide is added in each convolution operation to preserve the width and height of the data.
The {\bf BigNet} is also used to achieve higher accuracy.
The training epochs, learning rates and data augmentation settings remain the same as in the {\bf MNIST} experiments.

We evaluate DisturbLabel with the noise level ${\alpha}={20\%}$, and summarize the results in Table~\ref{Tab:ComparisonDigit}.
We can observe that DisturbLabel improves the recognition accuracy, either with or without using Dropout.
With data augmentation and both regularization methods, we achieve a competitive $2.02\%$ error rate.

\subsection{The CIFAR Datasets}
\label{Experiments:CIFAR}

\begin{table*}
\centering
\begin{tabular}{|l|r|r|||l|r|r|}
\hline
{\bf CIFAR10}                                    & w/o DA  & with DA  &
{\bf CIFAR100}                                   & w/o DA  & with DA  \\
\hline\hline
Zeiler~\cite{Zeiler_2013_Stochastic}             & $15.13$ & $-    $  &
Zeiler~\cite{Zeiler_2013_Stochastic}             & $42.51$ & $-    $  \\
Goodfellow~\cite{Goodfellow_2013_Maxout}         & $11.68$ & $ 9.38$  &
Goodfellow~\cite{Goodfellow_2013_Maxout}         & $38.57$ & $-    $  \\
Lin~\cite{Lin_2014_Network}                      & $10.41$ & $ 8.81$  &
Srivastava~\cite{Srivastava_2013_Discriminative} & $36.85$ & $-    $  \\
Wan~\cite{Wan_2013_Regularization}               & $-    $ & $ 9.32$  &
Lin~\cite{Lin_2014_Network}                      & $35.68$ & $-    $  \\
Lee~\cite{Lee_2015_Deeply}                       & $ 9.69$ & $ 7.97$  &
Lee~\cite{Lee_2015_Deeply}                       & $34.57$ & $-    $  \\
Liang~\cite{Liang_2015_Recurrent}                & $ 8.69$ & $ 7.09$  &
Liang~\cite{Liang_2015_Recurrent}                & $31.75$ & $-    $  \\
\hline\hline
{\bf LeNet}, no Reg.                             & $22.50$ & $15.76$  &
{\bf LeNet}, no Reg.                             & $56.72$ & $43.31$  \\
{\bf LeNet}, Dropout                             & $19.42$ & $14.24$  &
{\bf LeNet}, Dropout                             & $49.08$ & $41.28$  \\
{\bf LeNet}, DisturbLabel                        & $20.26$ & $14.48$  &
{\bf LeNet}, DisturbLabel                        & $51.83$ & $41.84$  \\
{\bf LeNet}, both Reg.                           & $19.18$ & $13.98$  &
{\bf LeNet}, both Reg.                           & $48.72$ & $40.98$  \\
\hline
{\bf BigNet}, no Reg.                            & $11.23$ & $ 9.29$  &
{\bf BigNet}, no Reg.                            & $39.54$ & $33.59$  \\
{\bf BigNet}, Dropout                            & $ 9.69$ & $ 7.08$  &
{\bf BigNet}, Dropout                            & $33.30$ & $27.05$  \\
{\bf BigNet}, DisturbLabel                       & $ 9.82$ & $ 7.93$  &
{\bf BigNet}, DisturbLabel                       & $34.81$ & $28.39$  \\
{\bf BigNet}, both Reg.                          & $ 9.45$ & $ 6.98$  &
{\bf BigNet}, both Reg.                          & $32.99$ & $26.63$  \\
\hline
\end{tabular}
\caption{
    Recognition error rates ($\%$) on the {\bf CIFAR10} and {\bf CIFAR100} datasets.
    DA: data augmentation (random cropping and flipping).
}
\label{Tab:ComparisonNatural}
\end{table*}

The {\bf CIFAR10} and {\bf CIFAR100} datasets~\cite{Krizhevsky_2009_Learning}
are both subsets drawn from the $80$-million tiny image database~\cite{Torralba_2008_80}.
There are $50000$ images for training, and $10000$ images for testing, all of them are $32\times32$ RGB images.
{\bf CIFAR10} contains $10$ basic categories, and {\bf CIFAR100} divides each of them into a finer level.
In both datasets, training and testing images are uniformly distributed over all the categories.
We use exactly the same network configuration as in the {\bf SVHN} experiments,
and add left-right image flipping into data augmentation with the probability $50\%$.

We evaluate DisturbLabel with the noise level ${\alpha}={10\%}$.
In {\bf CIFAR-100}, we slightly modify DisturbLabel by only allowing disturbing the label among $10$ finer-level categories.
We compare our results with the state-of-the-arts in Table~\ref{Tab:ComparisonNatural}.
Once again, DisturbLabel produces consistent accuracy gain in every single case, either with or without Dropout.
On {\bf CIFAR10}, the {\bf BigNet} with Dropout produces an excellent baseline ($7.08\%$ error rate),
and DisturbLabel further improves the performance ($6.98\%$ error rate) with a complementary regularization function to Dropout.
The success on the {\bf CIFAR} datasets verifies that DisturbLabel is generalized:
it not only works well in relatively simple digit recognition, but also helps natural image recognition tasks.

\subsection{The ImageNet Database}
\label{Experiments:ImageNet}

\begin{figure}
\begin{center}
    \includegraphics[width=\scatterwidth]{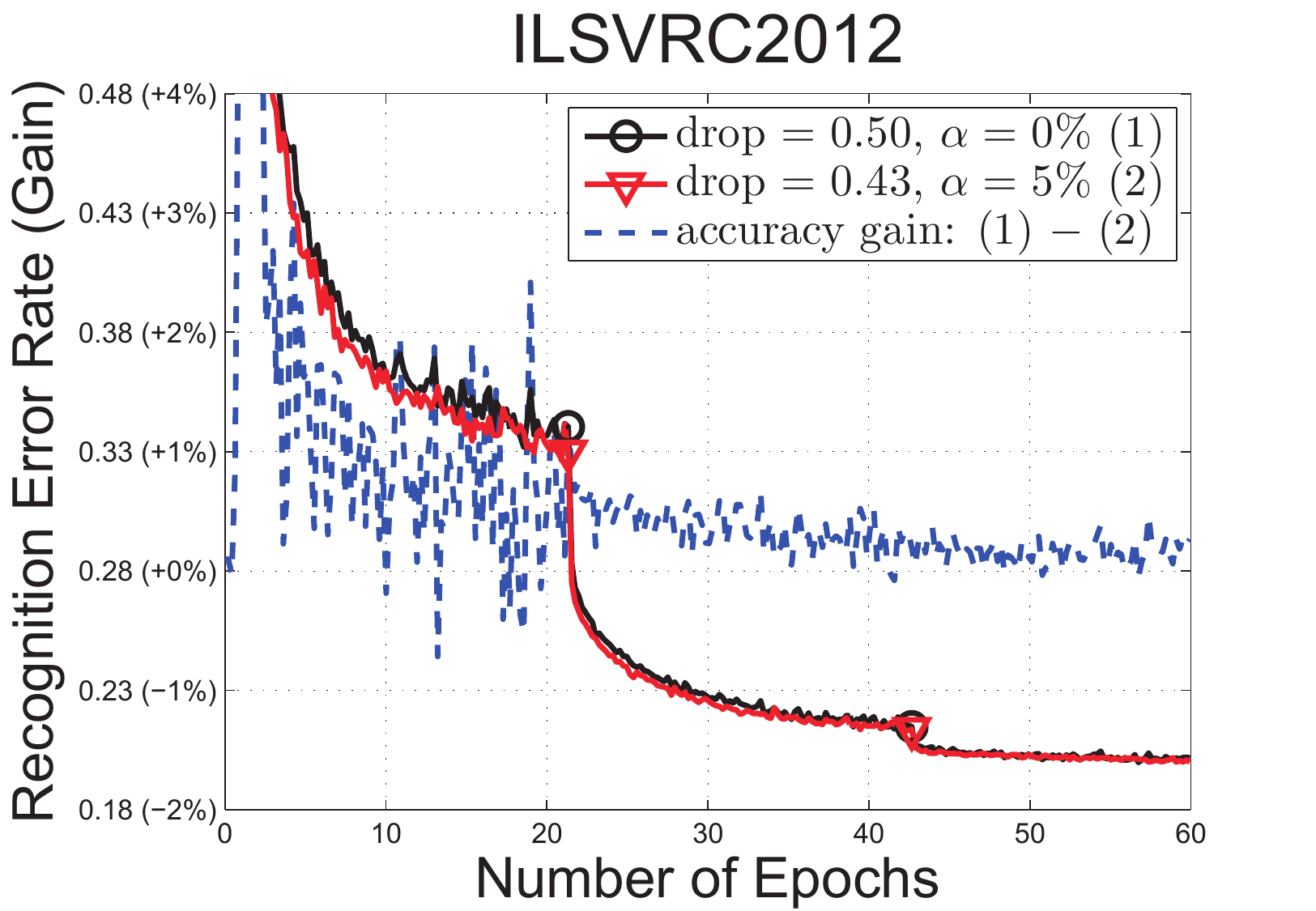}
\end{center}
\caption{
    The error rate and accuracy gain curves on the {\bf ILSVRC2012} validation set
    (the {\bf AlexNet}~\cite{Krizhevsky_2012_ImageNet} is used).
}
\label{Fig:ImageNet}
\end{figure}

Finally we evaluate DisturbLabel on the {\bf ILSVRC2012} classification task~\cite{Russakovsky_2015_ImageNet},
a subset of the {\bf ImageNet} database~\cite{Deng_2009_ImageNet} which contains $1000$ object categories.
The training set, validation set and testing set contain $1.3\mathrm{M}$, $50\mathrm{K}$ and $150\mathrm{K}$ images, respectively.
We use the {\bf AlexNet}~\cite{Krizhevsky_2012_ImageNet} (provided by the {\bf CAFFE} library~\cite{Jia_2014_CAFFE})
with the dropout rate $0.5$ as the baseline.
The {\bf AlexNet} structure is abbreviated as:
C$11$(S$4$)@$96$-MP$3$(S$2$)-LRN-C$5$(S$1$P$2$)@$256$-MP$3$(S$2$)-LRN-
C$3$(S$1$P$1$)@$384$-C$3$(S$1$P$1$)@$384$-C$3$(S$1$P$1$)@$256$-MP$3$(S$2$)-FC$4096$-D$0.5$-FC$4096$-D$0.5$-FC$1000$.
We note that when a training sample is chosen to be disturbed,
the label will be uniformly distributed among all the $1000$ classes,
introducing strong noises to the training process, even when the noise level is relatively low (${\alpha}={5\%}$ is used).
Therefore, we decrease the dropout rate to $0.43$ (less data are dropped) to perform weaker regularization.

The top-$1$ and top-$5$ error rates produced by the original {\bf AlexNet} are $43.1\%$ and $19.9\%$, respectively.
When DisturbLabel is adopted, the error rates are reduced to $42.8\%$ and $19.7\%$, respectively.
We emphasize that the accuracy gain is not so small as it seems
({\em e.g.}, the {\bf VGGNet}~\cite{Simonyan_2015_Very} combines two individually trained nets to get a $0.1\%$ gain),
which, once again, verifies that DisturbLabel and Dropout cooperate well to provide regularization in different aspects.
Figure~\ref{Fig:ImageNet} shows the error rate curve on the validation set.
After about $20$ epochs, the model with DisturbLabel produces higher recognition accuracy at each testing phase.

While we only evaluate DisturbLabel on the {\bf AlexNet},
we believe that it can also cooperate with other network architectures,
such as the {\bf GoogLeNet}~\cite{Szegedy_2015_Going} and the {\bf VGGNet}~\cite{Simonyan_2015_Very},
since regularization is a common requirement of deep neural networks.

\section{Conclusions}
\label{Conclusions}

In this paper, we present {\bf DisturbLabel}, a novel algorithm which regularizes CNNs on the {\bf loss layer}.
DisturbLabel is surprisingly simple, which works by randomly choosing a small subset of training data,
and intentionally setting their ground-truth labels to be incorrect.
We show that DisturbLabel consistently improves the network training process by preventing it from over-fitting,
and that DisturbLabel can be explained as an alternative solution of implicit model ensemble and data augmentation.
Meanwhile, DisturbLabel cooperates well with Dropout, which regularizes CNNs on the hidden neurons.
Experiments verify that DisturbLabel achieves competitive performance on several popular image classification benchmarks.

{\small
\bibliographystyle{ieee}
\bibliography{egbib}
}

\end{document}